\documentclass{article}

\usepackage[final]{corl_2025} 
\usepackage{graphicx}
\usepackage{caption}
\usepackage{subcaption}
\usepackage{amsmath}
\usepackage{amsfonts}
\usepackage{flushend}
\usepackage[x11names,table]{xcolor}
\usepackage{makecell}
\usepackage{diagbox}
\usepackage{booktabs}
\usepackage{pifont}
\newcommand{\cmark}{{\ding{51}}} 
\newcommand{\xmark}{{\ding{55}}}   
\usepackage[binary-units]{siunitx}

\title{SDS -- See it, Do it, Sorted: Quadruped Skill Synthesis\\from Single Video Demonstration}

\author{
  Maria Stamatopoulou$^{*}$ \quad
  Jeffrey Li$^{*}$ \quad
  Dimitrios Kanoulas \\
  Robot Perception Lab, University College London,  United Kingdom  \\
    {\small \texttt{\{maria.stamatopoulou.21, jeffrey.li.20, d.kanoulas\}@ucl.ac.uk}} 
}

\begin{document}
\maketitle
\begin{abstract}
Imagine a robot learning locomotion skills from any single video, without labels or reward engineering. We introduce \textbf{SDS} (``See it. Do it. Sorted.''), an automated pipeline for skill acquisition from unstructured demonstrations. Using GPT-4o, SDS applies novel prompting techniques, in the form of spatio-temporal grid-based visual encoding(\textbf{$G_{v}$}) and  structured input decomposition (\textbf{SUS}). These produce executable reward functions ($\mathcal{RF}$) from raw input videos. The $\mathcal{RF}$s are used to train PPO policies and are optimized through closed-loop evolution, using training footage and performance metrics as self-supervised signals.  SDS allows quadrupeds (e.g., Unitree Go1) to learn four gaits~--~\emph{trot}, \textit{bound}, \textit{pace}, and \textit{hop}~--~achieving 100\% gait matching fidelity, Dynamic Time Warping  (DTW) distance in the order of $10^{-6}$, and stable locomotion with zero failures, both in simulation and the real world. SDS generalizes to morphologically different quadrupeds (e.g., ANYmal) and outperforms prior work in data efficiency, training time and engineering effort. Further materials and the code are open-source under: \url{https://rpl-cs-ucl.github.io/SDSweb/}.
\end{abstract} 


\begin{figure}[htp!]
    \centering
    \includegraphics[height=5cm, width=0.9\linewidth]{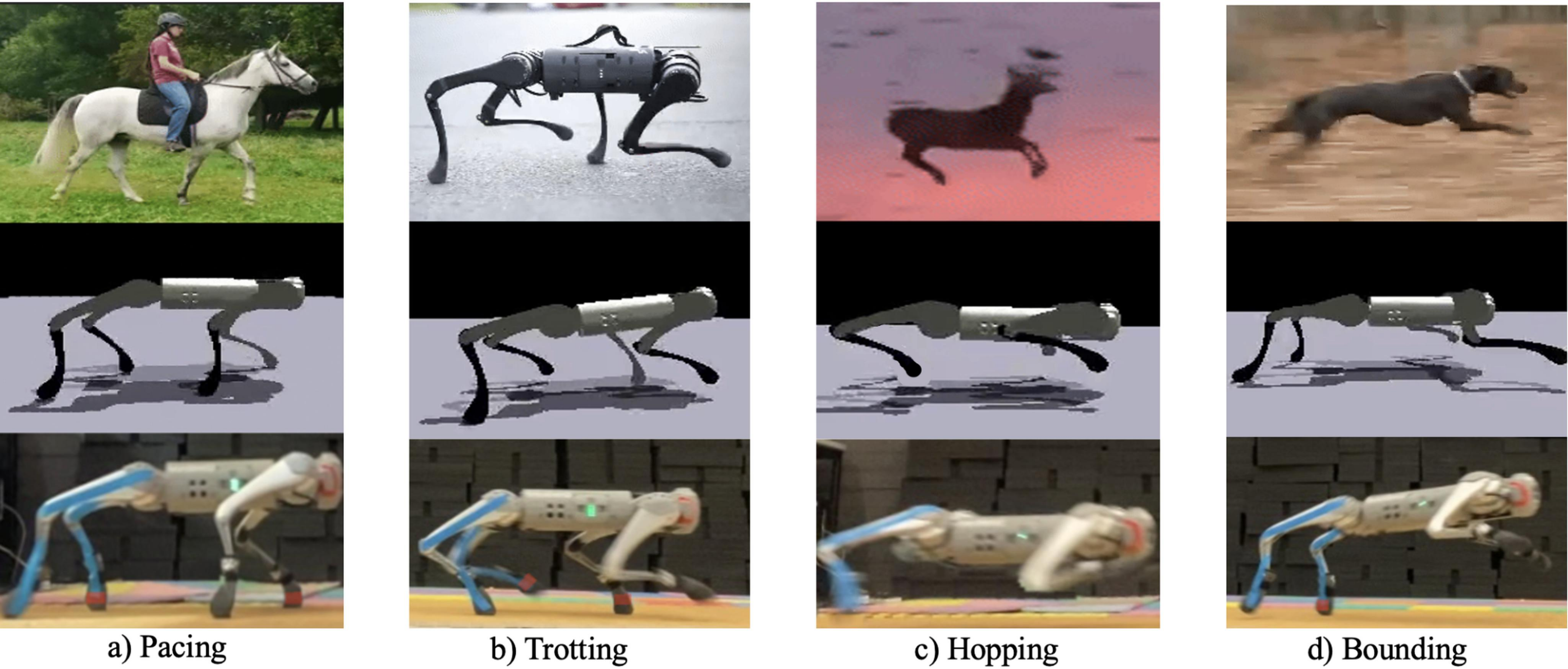}%
\caption{SDS's ability to imitate demonstrated skills (top), in simulation (center), and real-world (bottom). The blue tape corresponds to the rear legs, and the red tape to the left-side legs. The a)-d) ordering of the skills is assumed to be consistent throughout the paper.} 
    \label{fig:sds-evaluation}
\end{figure}
\section{INTRODUCTION}\label{sec:introuction}
Recent advances in large-scale foundational models—particularly language models (LLMs)\citep{dubey2024llama3herdmodels, openai2024gpt4technicalreport} and visual-language models (VLMs)\citep{lee2022uniclip, geng2023hiclip}—have unlocked new capabilities in robotics, enabling perception-action alignment and goal-driven reasoning ~\citep{stamatopoulou2024dippest, liu2024dipper}. While progress has largely focused on manipulation~\citep{wang2023mimicplay, sontakke2023roboclip}, quadrupedal locomotion remains a uniquely challenging domain due to its high-dimensional, contact-rich, and morphology-dependent dynamics~\citep{Kottege2022}. Acquiring diverse and robust gaits typically requires costly motion capture~\citep{Peng20, peng2020mocap}, heavily engineered reward functions~\citep{booth2023perils, rudin2021learning}, or manually tuned phase-scheduled controllers~\citep{Lee_2020}, making current pipelines brittle, non-scalable, and limited in skill diversity. While periodic in nature, they requiring precise, morphology-
specific tuning of phase offsets, contact timing, and stabilization gains
to differentiate across gaits and embodiments

To address this, we propose \textbf{SDS} (``See it. Do it. Sorted.''), an automated framework that learns diverse quadrupedal gaits directly from a single video, without MoCap, handcrafted rewards, or gait schedules. SDS leverages GPT-4o to synthesize executable reward functions ($\mathcal{RF}$) from video demonstrations via a novel multimodal prompting pipeline. Specifically, we introduce a grid-based video prompt ($G_v$) that adaptively samples motion-rich frames, and a chain-of-thought strategy (SUS) that decomposes high-level behaviors into semantically structured subgoals, enabling the VLM to generate compact, skill-differentiable reward terms. These rewards drive PPO-based policy learning in simulation and evolve through closed-loop visual feedback without predefined fitness metrics~\citep{ma2023eureka}. SDS explicitly targets periodic locomotion gaits, which remain an open challenge due to their morphology-specific coordination demands. Unlike existing pipelines requiring dense reward shaping and manual retuning, SDS produces transferable, class-separable gaits using only 2–6 reward terms and achieves zero-shot deployment across embodiments. Our key contributions are: \textbf{(a)} SDS, a general framework for quadrupedal gait learning from video; \textbf{(b)} a multimodal prompting system for semantically grounded reward synthesis; \textbf{(c)} sim-to-real deployment of four diverse gaits —pace, trot, hop, bound— on a Unitree Go1 robot.

\section{RELATED WORK}\label{sec:rw}
Learning from expert demonstrations for the imitation of skills has been explored extensively by the robotics community~\citep{shi2023waypoint, wang2023mimicplay, valassakis2022dome, totsila2023}. In the past, enabling robots to achieve locomotion required extensive reward engineering~\citep{rudin2021learning}. This is a challenging and time-consuming process, due to the sensitivity of deep learning algorithms to hyperparameters and weights, which requires trial-and-error fine-tuning, hence making it hard to develop even a single robot skill~\citep{Chen_2019}. 

As a response, robot control solutions shifted towards the dependence on domain experts and the creation of high-level action plans through learning-based approaches~\citep{yao2024local, chi2024diffusion}. MoCap systems are commonly used to extract expert data from animal movement in the form of keypoint locations (e.g., joints and base movement). Deep RL then maps these trajectories into low-level policies that replicate agile behaviors~\citep{Peng20, peng2020mocap, bohez2022imitate}. Although these methods outperform classical approaches by eliminating explicit environment modeling~\citep{kalashnikov2021mtopt}, they often suffer from reduced generalization and transferability to novel tasks and robotic platforms. Notable contributions leverage information from videos~\citep{yoon2024spatiotemporal}, by developing a spatio-temporal re-targeting framework that aligns the spatial trajectories and temporal dynamics of the source MoCap-recorded motion with the robot's capabilities and optimizes them using RL. Physics-informed 3D reconstruction from monocular videos further enhances motion learning~\citep{Zhang_SLOMO}. This method extracts keypoint trajectories, followed by offline trajectory optimization using contact-implicit constraints in a Model Predictive Controller (MPC). However, MPC's high computational cost and real-time optimization demands can introduce latency and require significant computational power.

Recent advances in generative AI have enabled more intuitive task description in robotics. Generative Adversarial Networks (GANs) have been explored to guide agent learning by dynamically selecting reference motions from expert datasets~\citep{peng2021amp,wang2021dynamic,ho2016generative}. However, mode collapse remains a key limitation due to the delicate balance between the generator and discriminator networks. Alternatively,~\citep{ma2023eureka} employs LLMs trained on code to automate the generation of RL reward functions, encoding task specifications for interpretation and refinement. However, the approach relies heavily on natural language descriptions and still requires manual, task-specific adjustments.
 We believe that ``a picture is worth a thousand words'', with such visual approaches introduced in~\citep{sontakke2023roboclip,yang2024}, where a Vision-Language Model (VLM) encodes visual and textual data in a shared embedding space, enabling one-shot robot policy learning. This eliminates manual reward engineering, though its generalization is currently limited to manipulator-based tasks and does not extend to mobile robots. Rocamonde et al.~\citep{rocamonde2024vlm} further use VLMs as zero-shot reward models for RL, providing natural language prompts to learn complex behaviors without manually defined rewards. However, the approach faces challenges such as limited spatial reasoning and sensitivity to visual realism in some environments.

\textbf{The SDS algorithm} is a novel automated pipeline that
eliminates the substantial human overhead associated with gait creation, by autonomously generating a set of RL reward functions ($\mathcal{RF}$) from a single demonstration video. It extends prior reward evolution methods~\citep{ma2023eureka} by removing the need for predefined fitness functions and MoCap data. Defining task fitness from video alone is challenging due to missing keypoints, spatial structure, and temporal cues, causing VLM-generated rewards to often miss critical behavioral aspects~\citep{rocamonde2024vlm}, eg., optimizing only forward motion may ignore gait-specific dynamics essential for imitation. SDS addresses this via a novel prompting strategy and a dynamic evaluation pipeline that iteratively refines $\mathcal{RF}$ using training feedback. We leverage GPT-4o, a vision-language model trained across diverse platforms, to improve generalization and enable fully autonomous reward generation, policy learning, and evaluation from a single video.

\section{METHOD}\label{sec:method}
\begin{figure}[htp!]
    \centering
\includegraphics[width=1\linewidth]{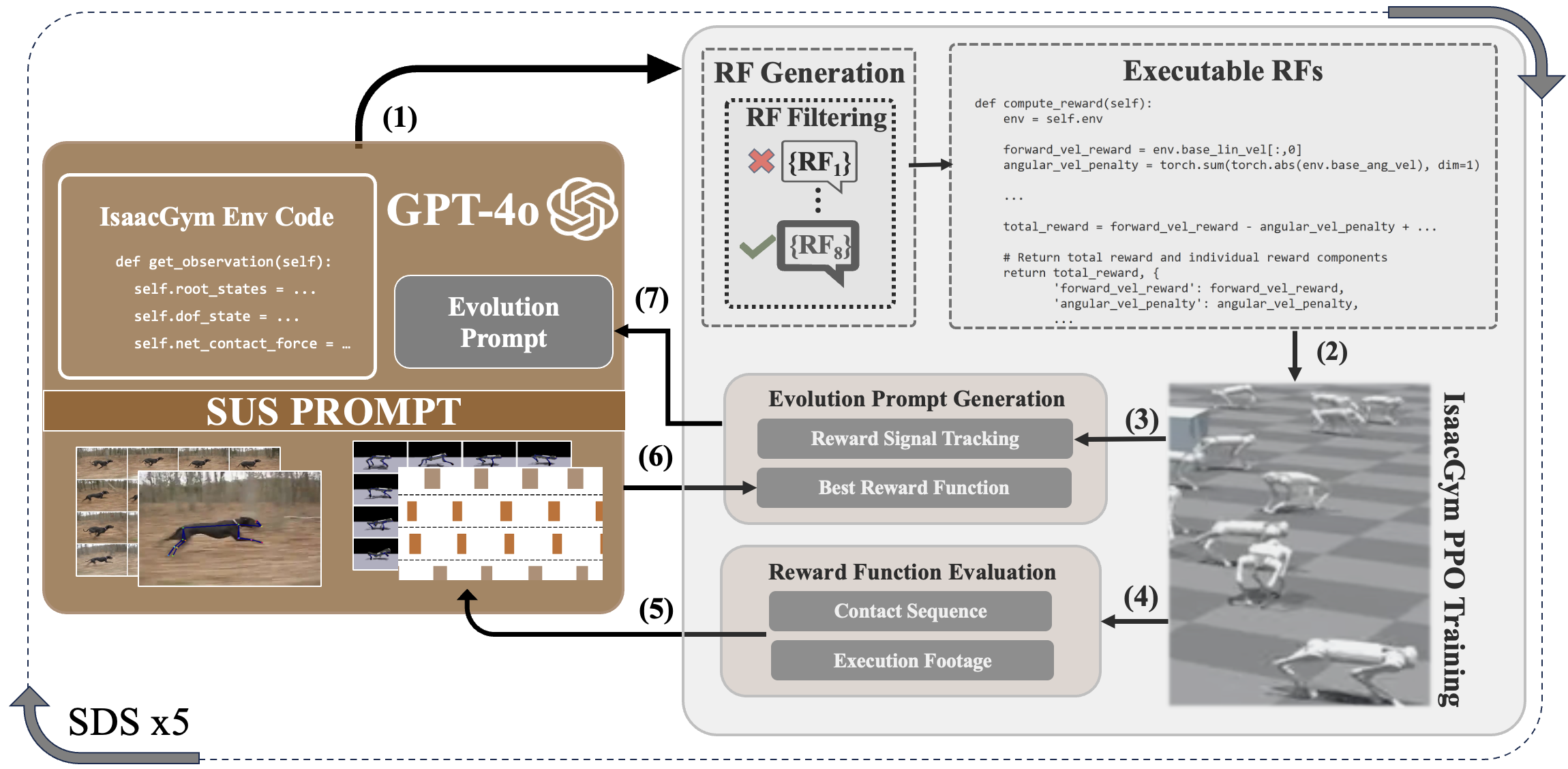}
    \caption{\textbf{SDS Method Overview.} The process begins by segmenting the demonstration video into $G_v$ and generating the SUS prompt. Each SDS iteration proceeds as follows: (1) GPT-4o generates a set $\mathcal{RF}$ of 8 candidate reward functions $RF_{i}$; (2) each executable $RF_{i}$ is used to train a PPO policy in IsaacGym; (3) sub-reward signals are monitored during training; (4) contact patterns ($CP$) and rollout footage ($G_s$) are recorded; (5) $RF_{i}$ performance is evaluated using the SUS prompt; (6) the best-performing reward $RF^{*}$ is selected; and (7) $RF^{*}$ is used to evolve the next iteration's $\mathcal{RF}$.}
    \label{fig:sds-method}
\end{figure}
SDS proposes a novel solution to generate executable RL pythonic reward functions ($\mathcal{RF}$), for quadrupedal robot skill learning, driven by a single video demonstration input. GPT-4o is selected as the VLM for its strong multimodal reasoning and generalization capabilities, with details in Ap.~\ref{sec:ap_vlmselection}. Our method introduces novel prompting techniques and a structured task fitness evaluation pipeline to enable the formulation of high-quality $\mathcal{RF}$ for high-fidelity skill imitation.

\subsection{Prompting Techniques}
\label{sec:method_prompting_techniques}
We develop novel prompting techniques for video input processing and high-fidelity task decomposition. 
\begin{figure}[htp!]
    \centering
    \includegraphics[height = 3.9cm, width=\linewidth]{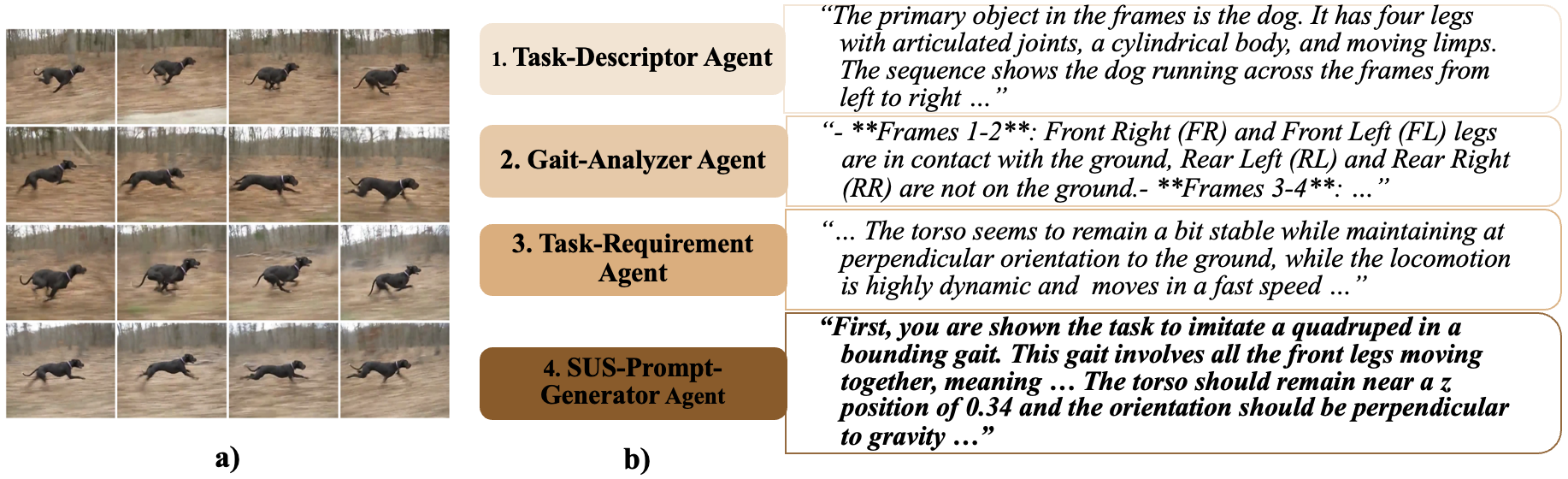}
    \caption{SDS Prompting Techniques for GPT-4o: a) Demonstration video frames, arranged in a grid ($\mathcal{G}_{v}$)  b) SUS skill decomposition into $4$ task-specific agents.}
    \label{fig:gridified}
\end{figure}

\begin{figure}[htp!]
    \centering
    \includegraphics[width=\linewidth]
    {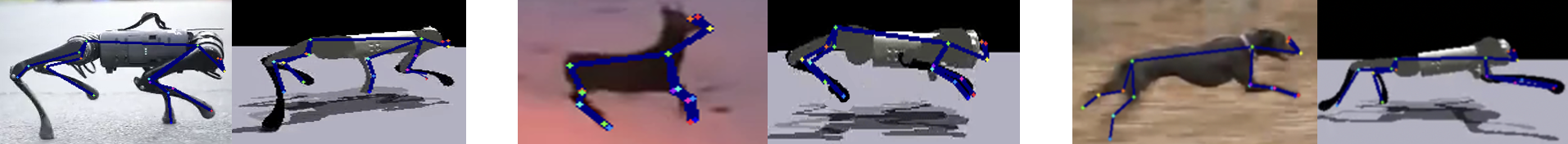}
    \caption{ViTPose++ Estimation on the demonstration frame and corresponding simulation frame.}
    \label{fig:pose_estimation}
\end{figure}

\paragraph{Grid-Frame Prompting:} GPT-4o lacks native video understanding and often struggles with temporal coherence. To efficiently standardize and encode video demonstrations for $\mathcal{RF}$ generation, we introduce a grid-based prompting method. Given a demonstration video $\mathcal{V} = \{I_1, I_2, \dots, I_T\}$ of duration $T$, we sample $n$ frames adaptively based on the velocity $v$ of the quadruped, with interval $\tau$:
\setlength{\abovedisplayskip}{0pt}
\[
\label{eq:video_promt}
n =\frac{T}{v}, \quad \tau = \frac{T}{n}, \tag{2}
\]
Lower velocities produce denser sampling to capture finer motion details~\citep{zhu2018}. The frames are arranged into a spatially uniform grid $\mathcal{G}_v \in \mathbb{R}^{h \times w}$, where $h = w = \sqrt{n}$. This strategy preserves temporal consistency, reduces token budget, and enables parallel visual processing. To mitigate hallucinations and misclassifications (e.g., often misclassifying a quadruped as a floating bench), we augment grid inputs with ViTPose++ keypoints~\citep{xu2023vitposevisiontransformergeneric}, providing a structured motion context, as shown in Fig.~\ref{fig:pose_estimation}. This augmentation improves GPT-4o prompting by mapping reward generation to the actual movement of the robot, reducing the ambiguity in the imitation of skills.

\paragraph{SUS Prompting:}
To ensure that the generated $\mathcal{RF}$ aligns with the demonstrated task, we introduce SUS ("See it. Understand it. Sorted."), a multistage prompting technique (Fig.~\ref{fig:gridified}.a). Inspired by cognitive reasoning~\citep{wei2023chainofthought, kojima2023large}, SUS aims to structure the information flow and improve interpretability and decision making, with prompts at Ap.~\ref{sec:ap_sds_prompts}.a. Given a $\mathcal{G}_{v}$, SUS uses a multi-agent framework to decompose the complex visual information into $4$ task-specific GPT-4o agents, by modifying each agent's system prompts. The \emph{a) Task-Descriptor Agent} is prompted to describe the most likely task being demonstrated. This information is transmitted to \emph{b) Gait-Analyzer Agent}, which analyzes the contact sequence and possible regular gait patterns. Next, \emph{c) Task-Requirement Agent} identifies additional key task characteristics to replicate the demonstration successfully. Finally, \emph{d) SUS-Prompt-Generator Agent} compiles all the information gathered to generate the final SUS prompt, used for the sampling of $\mathcal{RF}$.

\subsection{SDS Training Pipeline:}
\label{sec:learning}

The SDS training pipeline is structured into $7$ steps, corresponding to the numbers in Fig.~\ref{fig:sds-method}:

\textbf{(1) Reward Function ($\mathcal{RF}$) Generation:}
\label{sec:method_rf_gen}
We frame $\mathcal{RF}$ generation as a conditional code synthesis problem: GPT-4o is prompted to produce a set of Python reward functions $\mathcal{RF}$, conditioned on $\mathcal{G}v$, the SUS prompt, and environment-specific code (see Fig.\ref{fig:sds-method}(1)). While SDS is simulator-agnostic, we adopt NVIDIA IsaacGym\citep{margolis2022rapidlocomotionreinforcementlearning} for direct comparison with prior work~\citep{ma2024dreureka}. IsaacGym provides access to environment observations, including base pose, velocities, joint states, and foot contact flags, forming the input space for $\mathcal{RF}$. At each SDS iteration, GPT-4o outputs a set $\mathcal{RF} = \{ RF_i \}_{i=1}^n$, where each $RF_i$ is a dictionary containing individual sub-rewards and the aggregated total reward. Sub-rewards are computed over environment variables such as Boolean foot contacts ($\mathbb{R}^4$), joint angles and velocities ($\mathbb{R}^{12}$), and task-specific quantities derived from reference motions.$\mathcal{RF}$ generation is conditioned using a structured system prompt specifying the expected code format, input observations, and reward structure (using Ap.~\ref{sec:ap_sds_prompts}.b), together with a user prompt that injects video-specific and task-level objectives (using Ap.~\ref{sec:ap_sds_prompts}.c). All $RF_i$ candidates undergo static and runtime validation, ensuring correct Python syntax, API compatibility, and dynamic importability, with invalid samples being discarded (using~\ref{sec:ap_sds_prompts}.d). The set of valid functions $\mathcal{RF}_{\text{valid}} \subseteq \mathcal{RF}$ initializes policy training. After empirical tuning, we set $n = 8$ to balance candidate diversity with the risk of hallucinated or non-executable code.

\textbf{(2)-(4) Isaac Gym PPO Training:}
\label{sec:method_rf_train}
Each $RF_i \in \mathcal{RF}_{\text{valid}}$ defines an independent PPO  training run within IsaacGym- corresponding to an independent policy, using $4000$ parallel quadruped agents per environment. Policies are optimized over $1000$ iterations, producing $\mathbb{R}^{12}$ action vectors corresponding to target joint positions. Scalar sub-reward components returned by $RF_i$ are logged every $100$ iterations to track their contribution and temporal dynamics. Feasibility is enforced via PhysX, 25Nm torque clipping, and normalized rewards. Components showing unbounded growth are adaptively rescaled, while zero-gradient terms are flagged as uninformative and may be discarded in future iterations, ensuring that only meaningful sub-rewards are retained. 

\textbf{(5)-(6) Reward Function Evaluation:}
\label{sec:method_eval_step}
At each SDS iteration, the set $\mathcal{RF}_{\text{valid}}$ is evaluated post-training to identify the best-performing candidate $RF^*$ out of all the $RF_i$ policies. For each $RF_i \in \mathcal{RF}_{\text{valid}}$, the associated policy is deployed in simulation for a $1000$-timestep rollout. The resulting behavior is processed with ViTPose++ to extract dense keypoint trajectories, rendered into a temporally ordered image grid $\mathcal{G}_s$ (see Ap.~\ref{sec:ap_videoprompting}, Fig.~\ref{fig:ap_framefootage}), following the same procedure as in Step (1). In parallel, a binary contact sequence $\mathbb{R}^{4 \times 1000}$ is recorded, capturing per-timestep foot-ground contact states (using Ap.~\ref{sec:ap_sds_prompts}.d) and rendered as a contact plot ($CP$) visualizing inter-limb coordination over time. The combination of pose and contact diagnostics produces a multimodal behavioral trace that is richer and more interpretable than scalar rewards alone. For evaluation, a structured chat-style prompt is constructed comprising from $\mathcal{G}_s$, $CP$, and $\mathcal{G}_v$. These elements are embedded into a unified query, paired with a fixed system message template (\ref{sec:ap_sds_prompts}.f), which instructs GPT-4o to assess behavior according to task-specific criteria including \emph{postural stability}, \emph{gait periodicity}, and \emph{trajectory adherence}. GPT-4o returns a score vector, e.g., $[7, 8, 9]$, with scores ranging from $0-10$. The score vector is parsed into numerical form, summed, and used as the rollout’s aggregate performance metric. The $RF_i$ associated with the highest aggregate score is selected as the $RF^*$ for the next iteration. SDS with GPT-4o’s multimodal reasoning, replaces brittle hand-crafted rewards with behavior-aligned ones.

\textbf{(7) Reward Function Evolution:}
\label{sec:method_rf_evol}
Following evaluation, $RF^*$ is used to seed the next generation of $\mathcal{RF}$ candidates for the subsequent SDS iteration. Its refinement is guided by a structured prompt update that incorporates both the ${RF}^*$ code and feedback extracted from PPO training logs, in the form of scalar statistics for each sub-reward component converted into natural language summaries (using Ap.~\ref{sec:ap_sds_prompts}.g). If training succeeds, positive reinforcement prompts are added; otherwise, traceback diagnostics are included to address execution errors (using Ap.~\ref{sec:ap_sds_prompts}.h–i). This feedback, along with $\mathcal{G}_s$, $CP$, $\mathcal{G}_v$—is compiled into an updated user message. The system prompt (using Ap.~\ref{sec:ap_sds_prompts}.b) is reused to form a conversational context that conditions GPT-4o on both the prior $RF^*$ and its empirical performance and generates the new $\mathcal{RF}s$ set for the next SDS iteration. This iterative process enables gradient-free reward optimization via closed-loop GPT-4o interaction, progressively improving reward function quality through behavior-grounded and simulation-informed refinement.

\section{EXPERIMENTS \& RESULTS}
\label{sec:results}
SDS is evaluated on $4$ visually similar quadrupedal gaits of increasing dynamic complexity, providing a strong benchmark for imitation fidelity, with task details presented in Tab.~\ref{tab:skill_dynamics}. For each skill, we run $5$ SDS iterations using the learning parameters listed in Ap.~\ref{sec:ap_learning_parameters}, training within $1$ day using an NVIDIA RTX-4090 GPU. We transfer our SDS policy running at \SI{50}{Hz} zero-shot onto a Unitree Go1. The size of $\mathcal{G}_v$ for each skill is defined according to Eq.~\ref{eq:video_promt}, where the quadruped’s velocity is estimated by averaging ViTPose++ keypoints displacement across frames. The resulting $\mathcal{G}_v$ are depicted in Ap.~\ref{sec:ap_videoprompting}. The emergent sub-reward components of the final $RF^{*}$ used to train each skill policy are detailed in Ap.~\ref{sec:ap_rf_composition}, with specific sub-reward scores provided in Ap.~\ref{sec:ap_rw_scores}. 
\setlength{\abovedisplayskip}{0pt}
\begin{table}[htp!]
    \centering
    \renewcommand{\arraystretch}{0.95}
    \small
    \begin{tabular}{l|p{4.4cm}|p{3.6cm}|c|c}
    \toprule
    \textbf{Skill} & \textbf{Description} & \textbf{Emergent RF (Ap.~\ref{sec:ap_rf_composition})} & \textbf{Vel (m/s)} & $\boldsymbol{\mathcal{G}_v}$ \\
    \midrule
    Pace~\cite{horse_pace}  & Sync movement of adjacent limbs. & Vel, BH, Or, CP, AS, DoFL & 0.2 & $\mathbb{R}^{6 \times 6}$ \\
    Trot~\cite{unitree-go1}  & Sync movement of diagonal limbs. & Vel, BH, Or, LS, AS, DoFL & 0.5 & $\mathbb{R}^{4 \times 4}$ \\
    Hop~\cite{deer-hop}   & Sync movement of all limbs. & FM, DoFL & 1.2 & $\mathbb{R}^{4 \times 4}$ \\
    Bound~\cite{dog-run} & Sync movement of front limbs. & FM, BH, Or & 2.1 & $\mathbb{R}^{4 \times 4}$ \\
    \bottomrule
    \end{tabular}
    \caption{Overview of the demonstration skills, showing specific behaviors, emergent $RF^{*}$ sub-reward components, nominal velocities, and $\mathcal{G}_v$ sizes. Acronyms: Vel–Velocity, FM–Forward Motion, BH–Base Height, Or–Orientation, CP–Contact Pattern, LS–Limb Synchronization, AS–Action Smoothness, and DoFL–Degree of Freedom Limits.}
    \label{tab:skill_dynamics}
    \end{table}

\vspace{-3mm}  
\subsection{Skill Learning Evaluation}
To validate SDS’s skill imitation capabilities, we evaluate visual correspondence, contact sequence similarity, and locomotion stability of the learned policies, recording data from ten 1-minute runs. Evaluation is conducted both in simulation and on hardware, with imitation results shown in Fig.~\ref{fig:sds-evaluation}. 

\textbf{Task Imitation Evaluation:}
\setlength{\abovedisplayskip}{0pt}
\begin{table}[ht]
\centering
\renewcommand{\arraystretch}{0.95}
\small
\begin{tabular}{l|cccc||cccc}
\toprule
& \multicolumn{4}{c}{\textbf{a) DTW ($10^{-6}$)}} & \multicolumn{4}{c}{\textbf{b) Contact Sequences (\%)}} \\
\
\textbf{Skill} & sGo1 & rGo1& \textbf{Avg$_{go1}$} & sANYmal & sGo1 & {rGo1} & \textbf{Avg$_{go1}$} & sANYmal \\
\midrule
Pace  & 1.92 & 2.05 & 1.99 & - & 100 & 100 & 100 & - \\
Trot  & 1.28 & 2.44 & 1.86 & 151.49 & 100 & 100 & 100 & 100 \\
Hop   & 1.47 & 2.56 & 2.01 & - & 100 & 100 & 100 & - \\
Bound & 2.85 & 3.21 & 2.85 & 162.54 & 100 & 96  & 98  & 91.3 \\
\bottomrule
\end{tabular}
\caption{Average DTW distances and contact sequence matching between demonstration and learned skills over $1$-min trial $\times10$ runs (s-simulation, r-real-world).}
\label{tab:dtw}
\end{table}

\vspace{-0.4cm}

Dynamic Time Warping (DTW)~\citep{Müller2007} is employed to quantify frame-wise similarity between the demonstration footage and the SDS-learned policy, compensating for temporal misalignment by adapting time indices to minimize trajectory distance. Keypoint sequences extracted through ViTPose++, are spatially aligned via Iterative Closest Point (ICP) to correct global positional offsets. DTW analysis is conducted across all skills, with results presented in Table~\ref{tab:dtw}.a. All skills demonstrate strong trajectory correspondence, with trajectory distance values on the order of $10^{-6}$. A higher DTW value is observed for the bounding skill, primarily due to tracking errors caused by motion blur during high-speed hind limb movements.

\begin{figure}[ht!]
    \centering
    \includegraphics[height=4.7cm,width=0.9\linewidth]{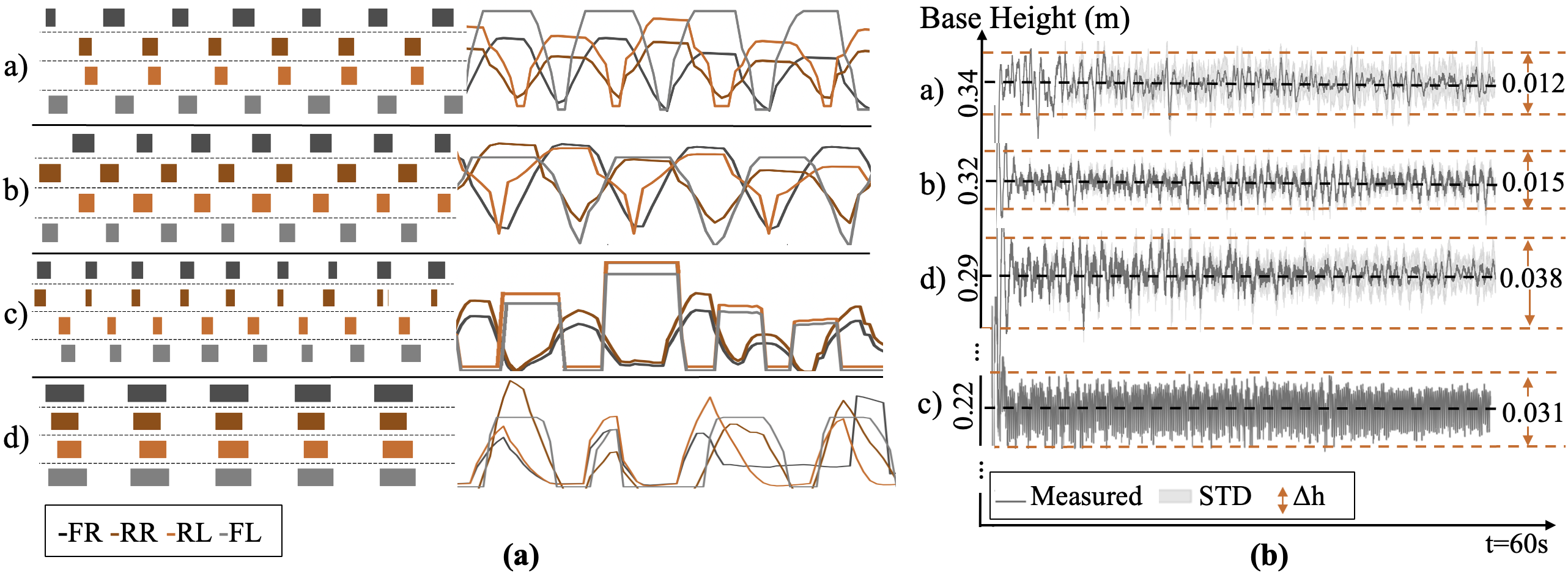}
    \caption{Gait evaluation results. (a) Contact sequences from simulation (right) and real-world smoothed force sensor readings (left) (F: Front, R: Rear; R: Right, L: Left). (b) Average base height and height fluctuation of the real robot in meters. (Note: graph not to scale.)}
    \label{fig:contacts}
\end{figure}

\textbf{Gait Imitation Evaluation:}
We further assess SDS's gait imitation fidelity (Fig.~\ref{fig:contacts}). Simulated contact sequences are encoded as Boolean foot-ground contact states over time, while real-world contact data are extracted from onboard force sensors and smoothed using a moving average filter~\citep{oppenheim2010discrete}. Average contact profiles are visualized in Fig.~\ref{fig:contacts}, and percent gait matching is reported in Table~\ref{tab:dtw}.b. Distinct gait patterns are evident across skills, with both simulated and real-world plots aligning with expected locomotion behaviors. All skills achieve 100\% contact sequence matching to the demonstrated pattern over 1-minute evaluation runs.

\vspace{-0.2cm}
\textbf{Locomotion Stability Evaluation:}
Real-world stability was evaluated by tracking base height fluctuations using both Phasespace~\cite{phasespace} motion capture (Fig.~\ref{fig:contacts}.b) and OpenCV-based object tracking (Ap.~\ref{sec:ap_real_world}). Across all skills, base height oscillations exhibited a low variance: pacing and trotting exhibited fluctuation variances of $3.6$ and $5.6\times10^{-5}$ m, respectively, while bounding and hopping showed slightly larger fluctuations of $2.2$ and $3.6\times10^{-4}$ m due to their dynamic nature. This demonstrates the stable control and robustness of SDS. Robustness was further validated by increasing ground friction via coarse socks, yielding no significant change (Wilcoxon test, $p = 0.07$). We compute the shortest distance from the robot CoM to the support polygon and the average angular velocity magnitude across roll, pitch, and yaw, extended with 50–110N lateral pushes applied for 2s at random intervals. The overall average StS across all skills and perturbations is very high - 1.77, given the maximum StS is 2. Further details are provided in Appendix~\ref{sec:ap_stability_score}. 

\subsection{Learning Generalization \& Component Ablations}
Ablations were conducted on the proposed prompting techniques ($\mathcal{G}{v}$, $SUS$) and the $\mathcal{RF}$ evolution mechanism components ($\mathcal{G}{s}$, $CP$), to evaluate their contributions to the method's performance. 

\begin{figure}[ht!]
    \centering
    \includegraphics[width=1\linewidth]{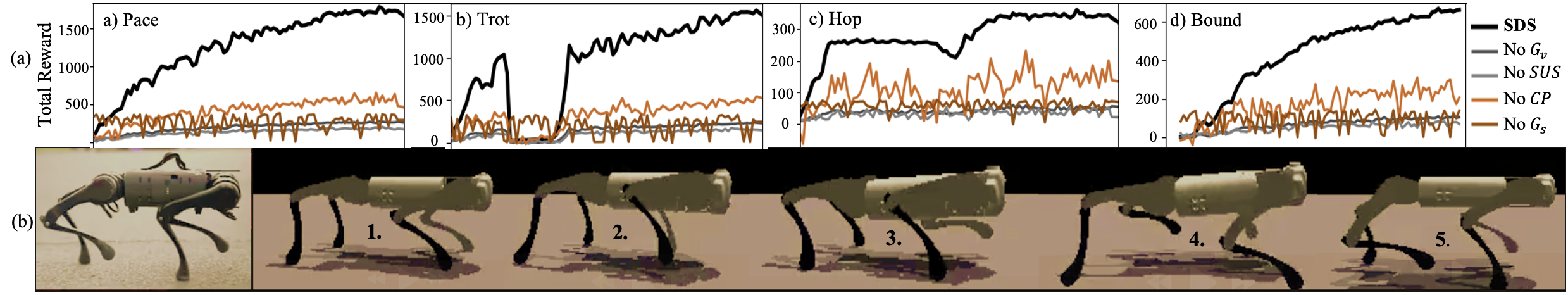}
    \caption{(a) Mean SDS reward signal and ablated variants, across $4$ skills trained on Unitree Go1. (b) Evolution of trotting behavior at a matched gait phase (T=$5$s) across the $5$ SDS reward iterations.}
    \label{fig:ablations}
\end{figure}

\textbf{$\mathcal{G}_{v}$ and $SUS$ Prompting:}
 Removing either prompting mechanism results in complete task learning failure (Fig.~\ref{fig:ablations}.a), with $0\%$ contact pattern alignment and DTW distances exceeding $\mathbf{150}$. These results underscore the necessity of both SUS and $\mathcal{G}_{v}$ prompting for scalable and temporally consistent skill acquisition. Without SUS, $\mathcal{RF}$ components fail to evolve meaningfully, remaining near-static across iterations and preventing task-specific adaptation. Without $\mathcal{G}{v}$, SDS must process frames independently, increasing computational cost by $\times\mathbf{16}$ due to slower convergence. Moreover, imitation fails because sequential processing lacks temporal coherence, which $\mathcal{G}{v}$ enforces by enabling spatially and temporally consistent reasoning across frames.
\newline
\textbf{$\mathcal{RF}$ Evolution:} Removing $\mathcal{G}_s$ eliminates qualitative validation, leading to unnatural movements, while removing $CP$ disrupts structured foot placement, causing drift and instability. Without both, task fidelity collapses, resulting in imitation and deployment failure. Ablating $\mathcal{G}_s$ or $CP$ reduces contact sequence matching to an average of $4\%$ and $19\%$ across skills, respectively. Removing $\mathcal{G}_s$ has a greater impact, as it provides global trajectory and motion structure critical for imitation, whereas $CP$ primarily enforces local foot contact patterns. Without $\mathcal{G}_s$, the agent loses overall motion fidelity, leading to failure even if some foot placements remain correct. Removing both components drives matching to $0\%$ and increases DTW distances beyond $100$, confirming their necessity for accurate skill acquisition. By examining the sub-components of the final $RF^{*}$ for each skill (Table~\ref{tab:skill_dynamics}), we can further interpret the effects of ablation. Skills such as Pace and Trot, which require precise limb synchronization, show strong reliance on $LS$ and $CP$ sub-reward components, with ablations causing significant reward degradation (Fig.~\ref{fig:ablations}.a). The Hop and Bound skills are less sensitive, relying more on $FM$ and $DoFL$ sub-reward components.The proposed evolution framework is further validated by the progressive alignment between $\mathcal{RF}$ and task behavior across SDS iterations ($1$–$5$), visualized for the trotting skill in Fig.\ref{fig:ablations}(b), and for all skills in Ap.~\ref{sec:ap_rw_evolution}.

\textbf{Generalization:} We evaluate SDS on the simulated ANYbotics ANYmal-D~\citep{anymal}, a quadruped that differs from the Unitree Go1 by being 38\,kg heavier, 25\,cm wider, 49\,cm taller, and featuring \textbf{inverted rear knee joints}. ANYmal-D is trained on trotting and bounding skills to cover both less and more dynamic gaits. Despite significant morphological differences, SDS achieves strong imitation performance, with an average DTW score of $157\times10^{-6}$, slightly higher than Go1 due to inverted joint keypoints, and a 95\% average contact sequence match. These results demonstrate successful skill transfer and generalization, with imitation pairs shown in Ap.~\ref{fig:anymal}.

\subsection{SDS compared with SOTA methods}
We compare SDS with works of similar scope~\citep{ma2023eureka,ma2024dreureka,sontakke2023roboclip,Zhang_SLOMO,rocamonde2024vlm}, with a summary presented in Table~\ref{tab:sds_comparison} and evaluation metrics detailed in Ap.~\ref{sec:ap_comaprison_metrics}.
Unlike~\citep{ma2023eureka}, SDS eliminates the need for manual task fitness design and enhances interpretability through the introduction of visual prompting. Compared to~\citep{sontakke2023roboclip}, SDS achieves superior quadrupedal skill imitation; the method in~\citep{sontakke2023roboclip} failed to reproduce demonstrations even after three additional training days, due to its reliance on a task-specific VLM designed for manipulation.
SDS also outperforms~\citep{Zhang_SLOMO}, which requires eight RTX 3080 GPUs and eight hours of optimization per one-minute demonstration, along with a continuous workstation connection. In contrast, SDS runs fully onboard and in real time.
Overall, SDS requires no human intervention, replicates motion directly from video, trains efficiently, generalizes through a general-purpose VLM, and supports fully onboard execution across all skills.
\begin{table}[ht]
\centering
\renewcommand{\arraystretch}{0.8}
\small
\begin{tabular}{l|cccccc}
\toprule
\textbf{Method/Metrics} & \textbf{Train Time} & \textbf{DTW}  & \textbf{No Human} & \textbf{No Input Extras} & \textbf{Real World} \\
\midrule
Eureka~\citep{ma2023eureka}           & 1 day   & $\infty$           & \xmark & \xmark & \xmark \\
DrEureka~\citep{ma2024dreureka}       & 1 day   & $>$10               & \cmark & \xmark & \cmark \\
RoboCLIP~\citep{sontakke2023roboclip} & 4 days  & $\infty$    & \cmark & \xmark & \xmark \\
SloMo~\citep{Zhang_SLOMO}             & 8 days  & $>$10          & \xmark & \xmark & \cmark \\
VLM-RM~\citep{rocamonde2024vlm}       & 3 days   & $\infty$        & \cmark & \xmark & \xmark \\
\textbf{SDS (Ours)}                   & 1 day   & 1.3$\times10^{-6}$             & \cmark & \cmark & \cmark \\
\bottomrule
\end{tabular}
\caption{Comparison of SDS with SOTA methods. Metrics include: Train Time for a single skill using a RTX-4090, DTW imitation accuracy for trotting skill, need for human intervention in reward design, extra task inputs, and real-world validation. More detail on the metrics in Ap.~\ref{sec:ap_comaprison_metrics}}
\label{tab:sds_comparison}
\end{table}
\vspace{-0.2cm}
\section{CONCLUSION}\label{sec:conclusion}
We present SDS, a pipeline for learning quadrupedal skills from a single demonstration video. SDS leverages GPT-4 to automatically generate reward functions ($\mathcal{RF}$), enabling PPO training in IsaacGym without manual reward engineering. Our key contributions include a novel prompting strategy and an autonomous $\mathcal{RF}$ evaluation and evolution framework, allowing precise capture of skill-specific dynamics. SDS was validated through extensive real-world experiments, achieving robust skill imitation across $4$ skills with zero resets and low-variance base height oscillations. Quantitatively, SDS attained $>96\%$ contact sequence matching with demonstrations and maintained DTW distance in the scale of $10^{-6}$ across all tasks. Compared to SOTA methods, SDS eliminates the need for manual fitness design while increasing fidelity, operates fully onboard and reduces training time.

\section{LIMITATIONS \& FUTURE WORK}
While SDS shows promising results in the imitation of single-video skills, we acknowledge several key limitations. The system currently relies on side-view demonstrations with clear limb visibility, limiting generalization to arbitrary viewpoints and occluded settings. Limited keypoint visibility in videos of short-legged or furry animals (e.g., Persian cats) led to reduced tracking accuracy and poorer imitation quality. We also observe lower imitation success rates for static or less dynamic skills, such as a horse rearing onto two legs. This may be due to limited temporal variation in the demonstration, reduced reliability of contact-based cues, and reduced feedback signal during PPO training. SDS is also validated only on flat terrain and implicitly assumes environmental similarity between simulation and demonstration. However, enhanced real-world deployment demands robustness to diverse terrain properties and domain shifts, which we plan to address through terrain-aware reward generation and domain randomization. Additionally, SDS assumes access to the full state of the simulator, which is not always feasible in real-world conditions; we intend to explore learning from onboard sensing alone using recurrent or belief-based policies. The current single-skill-per-SDS training approach restricts scalability, motivating ongoing work on hierarchical and skill-conditioned architectures for multi-skill integration. Further, SDS currently focuses on periodic locomotion skills, using contact sequences as a lightweight, interpretable feedback signal. We recognize that extending to non-periodic behaviors will require temporally grounded visual inputs (e.g., keypoint trajectories), modified SUS prompting, and new evaluation metrics. Finally, while SDS generalizes to a morphologically distinct quadruped (ANYmal), its effectiveness on platforms with entirely different morphology, such as humanoids, remains untested. We aim to expand  SDS to broader morphologies and to analyze the impact of structure and actuation on learning dynamics. \\
\textbf{Future work} aims to validate real-world policies in outdoor and unstructured environments, and to extend the skill set to include key locomotion behaviors such as climbing stairs, twisting, and other complex maneuvers. We then aim to combine all skills to enable multi-skill control within a single policy for adaptable quadruped locomotion across diverse environments. In the longer term, we plan to extend SDS to other mobile platforms, with a primary focus on humanoids, which exhibit higher degrees of freedom and pose greater control challenges. We also plan to extend SDS toward non-periodic and trajectory-conditioned skills using temporally-aware rewards (e.g., DTW, trajectory curvature) and integrate high-level modules - e.g., planning, obstacle avoidance~\citep{zhang2025,Linghong2024, Roth_2024}, while maintaining SDS’s focus on observable, deployable behaviors relevant to real-world quadrupedal autonomy.

\bibliography{vlm}  

\clearpage
\appendix
\section{Appendix}
This section provides additional information to support the main paper.

\subsection{VLM Selection:}\label{sec:ap_vlmselection}
Vision-Language Models (VLM) encode joint visual-text representations through large-scale multimodal training~\citep{bordes2024introductionvisionlanguagemodeling}. Given an input sequence $V = {I_1, I_2, ..., I_T}$ and textual descriptions $T$, parameters $\theta$ are optimized by masked modeling, contrastive objectives, or causal generation. Standard VLMs, trained primarily on image-text pairs, struggle with temporal reasoning essential for skill learning. Video-based adaptations extend pre-trained encoders $f_{\theta}(I)$ with temporal mechanisms~\citep{NEURIPS2022_381ceeae, ni2022expanding} or adopt end-to-end video pre-training with masked modeling~\citep{lin2022swinbert} and contrastive learning~\citep{geng2023hiclip}. We evaluated other SOTA VLMs~\citep{lin2023videollavalearningunitedvisual, tong2022videomaemaskedautoencodersdataefficient} under identical structured prompts which despite strong VQA performance, they failed to robustly produce executable Python code. We also observed that SOTA pipelines using morphology specific VLMs (e.g., S3D) limit cross-embodiment generalization. GPT-4o(ision)~\citep{openai2023gptv} was selected for its ability to synthesize structured, executable Python code from visual prompts and its broad multi-modal training, which is critical for extending SDS beyond quadrupeds. GPT-4o's robust visual reasoning and multi-modal dataset ($\mathcal{D}$) spanning both animal and human data, will aid in improving generalization of SDS. GPT-4o follows a two-stage training:
\setlength{\abovedisplayskip}{0.1pt} 
\begin{equation}
\mathcal{L}_{\text{GPT-4o}} = \mathbb{E}_{(x_t, x_v) \sim \mathcal{D}} \left[ \log P(x_t, x_v \mid \theta) \right]
\end{equation}
where $x_t$ and $x_v$ are textual and visual tokens, respectively. Fine-tuned for domain-specific applications, GPT-4o enables structured task decomposition and robust visual input reasoning, making it well-suited for robot skill learning.
\newpage
\subsection{SDS Prompting}
\subsubsection{Video Prompting} 
\label{sec:ap_videoprompting}
The grids ($G_{V})$ of the demonstration videos used to train the different SDS skills are presented in Fig.~\ref{fig:ap_videoprompting}. 
\begin{figure}[htp!]
    \centering
    \includegraphics[height=7.8cm,width=0.7\linewidth]{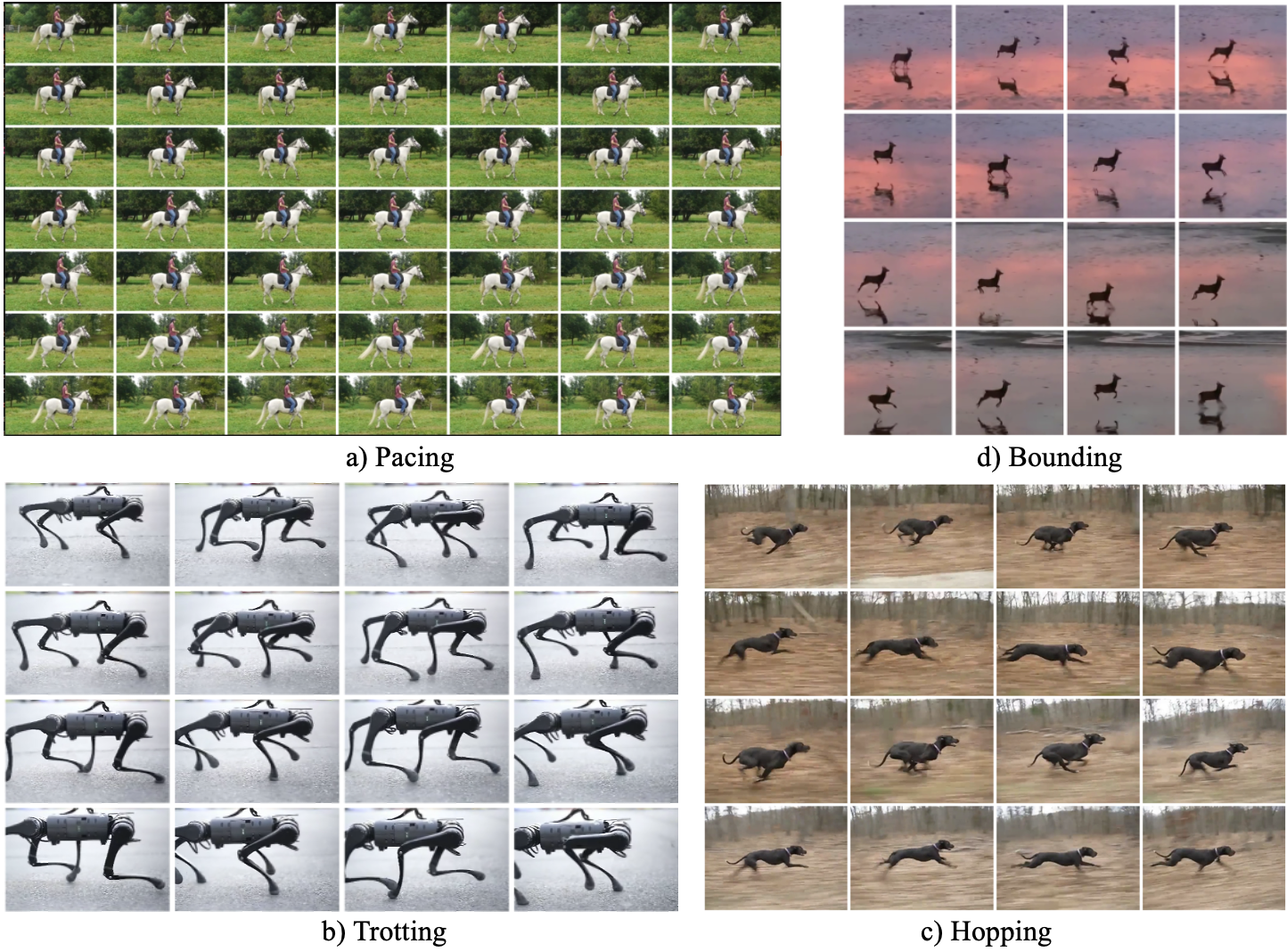}
    \caption{Demonstration videos arranged in a grid formation ($\mathcal{G}_{v}$), serving as input to GPT-4o for SDS processing.}
    \label{fig:ap_videoprompting}
\end{figure}

\begin{figure}[htp!]
    \centering
    \includegraphics[height=7cm,width=0.8\linewidth]{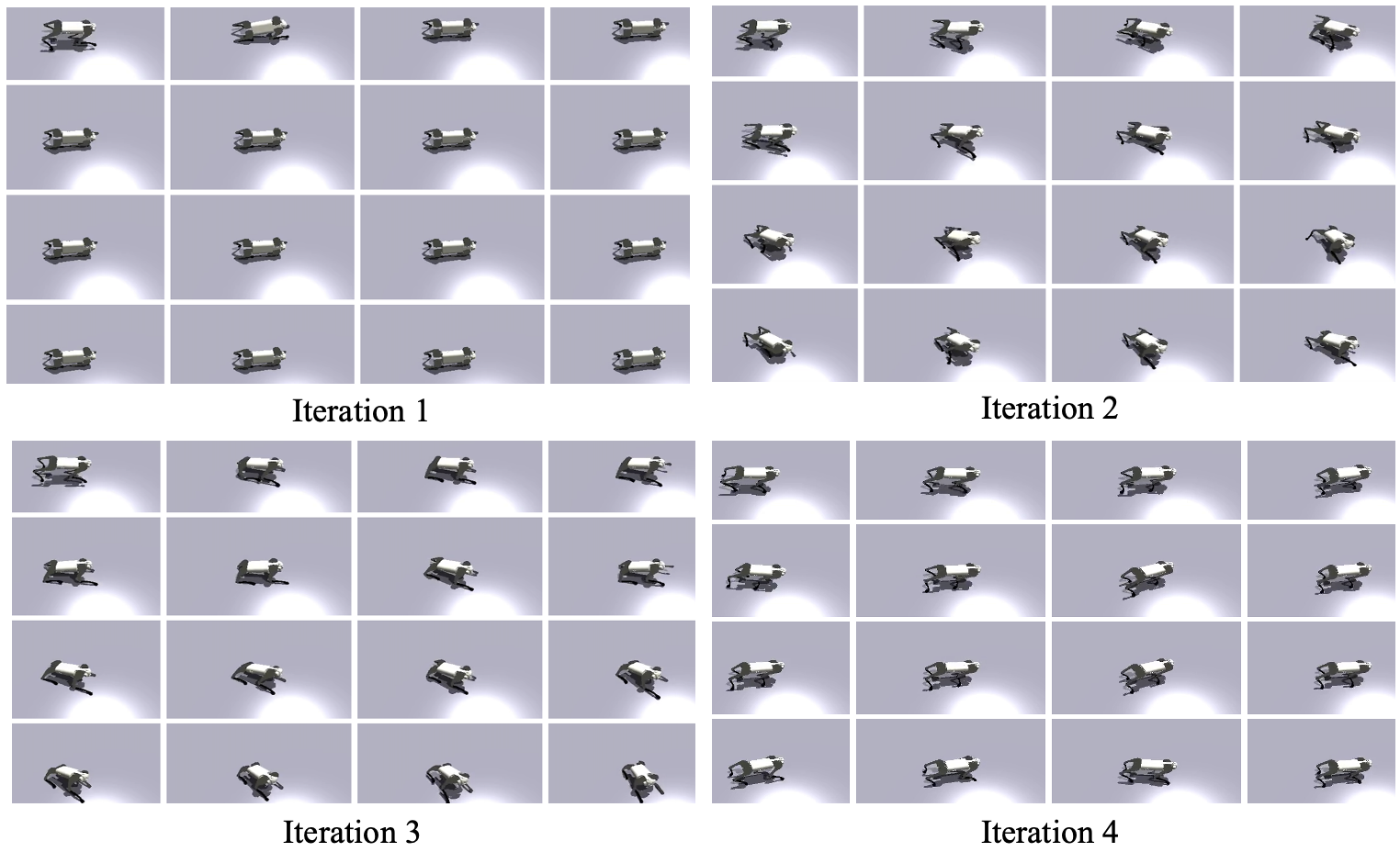}
    \caption{Simulation footage arranged in a grid formation ($\mathcal{G}_{s}$), serving as input as GPT-4o input.}
    \label{fig:ap_framefootage}
\end{figure}

The simulation footage frames for the Trotting skill across SDS iterations, organized into grids $\mathcal{G}_s$, for GPT-4 prompting, presented in Fig.~\ref{fig:ap_framefootage}.

\subsubsection{SUS Prompting}
\label{sec:sus_videoprompting}
Task decomposition of the trotting demonstration video produced by the four task-specific GPT-4o agents, presented in Fig.~\ref{fig:ap_videoprompting}.
\begin{figure}[htp!]
    \centering
    \includegraphics[height=4.2cm, width=\linewidth]{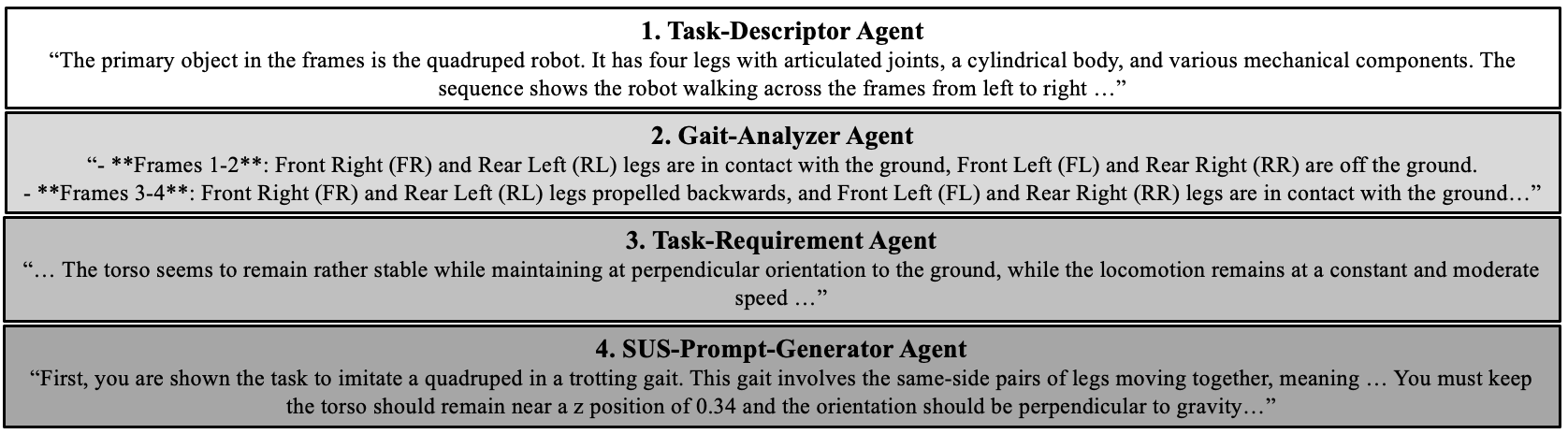}
    \caption{SUS Prompting:4 Task-Specific GPT-4o Agents decomposing the Trotting task.}
    \label{fig:ap_videoprompting}
\end{figure}

\subsubsection{Method Prompts}
\label{sec:ap_sds_prompts}
\begin{table}[htp!]
    \centering
    \begin{tabular}{p{5.0cm} p{0.5cm} p{7.2cm}}
        \toprule
        \textbf{Prompt File} & \textbf{Step} & \textbf{Description} \\
        \midrule
        \texttt{a) sus\_generator, SUS} & (0) & Prompt used to generate the structured SUS (Skill Understanding String) from gait annotations and demonstration metadata; output is inserted into the reward generation prompt. \\
        \texttt{b) init\_sds\_system} & (1) & System prompt instructing GPT-4o to generate executable reward functions compatible with Isaac Gym's observation API. Defines function structure and behavioral objectives. \\
        \texttt{c) sds\_user} & (1) & User prompt injecting task-specific goals based on gait analysis and demonstration video, guiding reward synthesis toward desired behaviors. \\
        \texttt{d) code\_output\_tip} & (1,7) & Auxiliary prompt appended to enforce complete, syntactically valid code outputs, discouraging truncation or malformed syntax. \\
        \texttt{e) contact\_sequence\_system} & (5) & Supplementary system prompt used when evaluating foot-ground contact sequences, emphasizing gait periodicity and limb coordination. \\
        \texttt{f) init\_task\_evaluator\_system} & (5) & System prompt defining multimodal evaluation criteria (stability, periodicity, trajectory fidelity) for GPT-4o-based policy scoring. \\
        \texttt{g) policy\_feedback} & (7) & Positive feedback prompt included when a training run completes successfully, summarizing reward statistics and learning progress. \\
        \texttt{h) code\_feedback} & (7) & Positive reinforcement prompt returned when generated reward code executes correctly and produces meaningful learning dynamics. \\
        \texttt{i) execution\_error\_feedback} & (7) & Diagnostic prompt containing traceback and error information, sent to GPT-4o when code fails to compile or execute during validation. \\
        \bottomrule
    \end{tabular}
    \caption{Prompt files used in SDS and their corresponding roles within each step of the methodology. Prompt files can be accessed at: \url{https://github.com/sdsreview/SDS_ANONYM/tree/main/SDS/prompts}} \label{tab:sds_prompts}
\end{table}
\subsection{Learning Parameters}\label{sec:ap_learning_parameters}
We present the learning parameters used to train SDS.
    \begin{table}[htp!]
        \centering
        \begin{tabular}{l|ccccccc}
        \toprule
        \textbf{LR} & \textbf{Clip} & \textbf{Entropy} & \textbf{Mini-batches} & \textbf{Epochs} & \textbf{Gamma} & \textbf{Max Iters} \\
        \midrule
        0.001 & 0.2 & 0.01 & 4 & 5 & 0.99 & 1500 \\
        \bottomrule
    \end{tabular}
\caption{PPO training hyperparameters used consistently across all SDS skills.}
\label{tab:training_config_shared}
\end{table}

\begin{table}[htp!]
    \centering
    \begin{tabular}{l|l}
        \toprule
        \textbf{Parameter} & \textbf{Value} \\
        \midrule
        Initialization Noise Std. & 1.0 \\
        Actor Hidden Layers & [512, 256, 128] \\
        Critic Hidden Layers & [512, 256, 128] \\
        Activation Function & ELU \\
        Adaptation Branch Hidden Dims & [[256, 32]] \\
        Env Factor Encoder Input Dims & [18] \\
        Env Factor Encoder Latent Dims & [18] \\
        Env Factor Encoder Hidden Dims & [[256, 128]] \\
        \bottomrule
    \end{tabular}
    \caption{Actor-Critic (AC) network configuration used across SDS skill training.}
    \label{tab:ac_network_config}
\end{table}
\subsection{Reward Functions}\label{sec:ap_rw_description}
\subsubsection{Reward Components Description}\label{sec:ap_rf_composition} Presenting the potential sub-rewards emerging within the ${RF_i}$ dictionary.
\begin{table}[htp!]
\centering
    \begin{tabular}{l|p{6cm}c|c}
        \toprule
        \textbf{Component} & \textbf{Description} & \textbf{Unit} & \textbf{Dim.} \\
        \midrule
        Velocity (Vel)  & Encourages matching the commanded base linear velocity. & [m/s] & $\mathbb{R}^{4 \times 4}$ \\
        Forward Motion (FM) & Rewards forward translational motion aligned with the command direction. & [m/s] & $\mathbb{R}^{4 \times 4}$ \\
        Base Height (BH) & Rewards maintaining a target base height to ensure locomotion stability. & [m] & $\mathbb{R}^{1}$ \\
        Orientation (Or) & Penalizes deviation from upright orientation using quaternion distance. & [rad] & $\mathbb{R}^{1}$ \\
        Contact Pattern (CP) & Encourages limb contact timings to match a desired gait pattern (e.g., pacing). & [binary match] & $\mathbb{R}^{4 \times T}$ \\
        Limb Sync (LS) & Rewards synchronized limb movement (e.g., diagonal or bounding gaits). & [unitless] & $\mathbb{R}^{1}$ \\
        Action Smoothness (AS) & Penalizes abrupt action changes to promote smoother joint torques. & [rad/s] & $\mathbb{R}^{n}$ \\
        DoF Limits (DoFL) & Penalizes joint positions near mechanical limits to avoid over-extension. & [rad] & $\mathbb{R}^{n}$ \\
        \bottomrule
    \end{tabular}
\caption{Descriptions, units, and dimensions of reward components used across different SDS skills.}
\label{tab:reward_descriptions}
\end{table}

\subsubsection{$RF^{*}$ Sub-Reward Scores}\label{sec:ap_rw_scores}
We present the sub-reward scores for the final policy of each skill in Table~\ref{tab:reward_components}, and the aggregate reward sum and failure rate for each skill in Table~\ref{tab:reward_and_resets}.
\begin{table}[htp!]
    \centering
    \begin{tabular}{l|rrrrrrrr}
        \toprule
        \textbf{Skill} & \textbf{Vel} & \textbf{FM} & \textbf{BH} & \textbf{Or} & \textbf{CP} & \textbf{LS} & \textbf{AS} & \textbf{DoFL} \\
        \midrule
        Pace  & 0.026 & --    & 11.191 & 0.001  & 4.140 & --    & 0.874 & 11.950 \\
        Trot  & 0.153 & --    & 11.192 & 0.001  & --    & 7.717 & 0.875 & 2.391  \\
        Hop   & --    & 1.480 & --     & --     & --    & --    & --    & 0.006  \\
        Bound & --    & 0.514 & 11.574 & 11.474 & --    & --    & --    & --    \\
        \bottomrule
    \end{tabular}
\caption{Final $RF^{*}$ sub-reward component values for each learned skill. Missing values (--) indicate the absence of that reward in the respective skill.}
\label{tab:reward_components}
\end{table}
Total $RF^{*}$ values for each skill, showing the total aggregate reward value and reset value- occurring when the robot reaches a termination condition (base or joint hit the ground, orientation diverges from  limits). 
\begin{table}[htp!]
    \centering
    \begin{tabular}{l|rr}
        \toprule
        \textbf{Skill} & \textbf{Total Reward} & \textbf{Reset Events} \\
        \midrule
        Pace  & 3.301  & 0 \\
        Trot  & 14.821 & 0 \\
        Hop   & 5.986 & 0 \\
        Bound & 9.157  & 0 \\
        \bottomrule
    \end{tabular}
\caption{Total reward and reset events for each learned skill. All policies remained stable, with no reset events during training.}
\label{tab:reward_and_resets}
\end{table}
\subsubsection{Policy Stability Evaluation}\label{sec:ap_stability_score}
We further evaluate robustness using two stability metrics: the shortest distance from the center of mass to the support polygon (CoM\textsubscript{dist}) and the average angular velocity magnitude ($|\omega|$) computed over roll, pitch, and yaw. These are combined into a single score, the Stability-to-Speed ratio (StS), defined as:
\begin{equation}
\text{StS} = 2 - \left[ \text{clip}(\text{CoM}_{\text{dist}},\ 0,\ 1) + \text{clip}(|\omega|,\ 0,\ 1) \right]
\end{equation}

A higher StS indicates better stability, with a maximum possible score of 2.
To further test robustness, we applied lateral perturbation forces ranging from 50N to 110N for 2s at random time intervals. We report StS scores under both unperturbed (0N) and maximum perturbation (110N) conditions.

\begin{table}[h]
\centering
\caption{Policy Stability Score (StS) under 0N and 110N lateral perturbations (2s duration).}
\begin{tabular}{lcc|c}
\toprule
\textbf{Gait} & \textbf{StS (0N)} & \textbf{StS (110N)} & \textbf{Skill Mean} \\
\midrule
Pace  & 1.86 & 1.75 & 1.81 \\
Trot  & 1.92 & 1.79 & 1.86 \\
Hop   & 1.77 & 1.64 & 1.71 \\
Bound & 1.78 & 1.68 & 1.73 \\
\midrule
\textbf{Perturbation Mean} & \textbf{1.83} & \textbf{1.72} & \textbf{1.77} \\
\bottomrule
\end{tabular}
\end{table}
\subsubsection{$\mathcal{RF}$ Evolution}\label{sec:ap_rw_evolution}
\begin{figure}[htp!]
    \centering
    \includegraphics[width=1\linewidth]{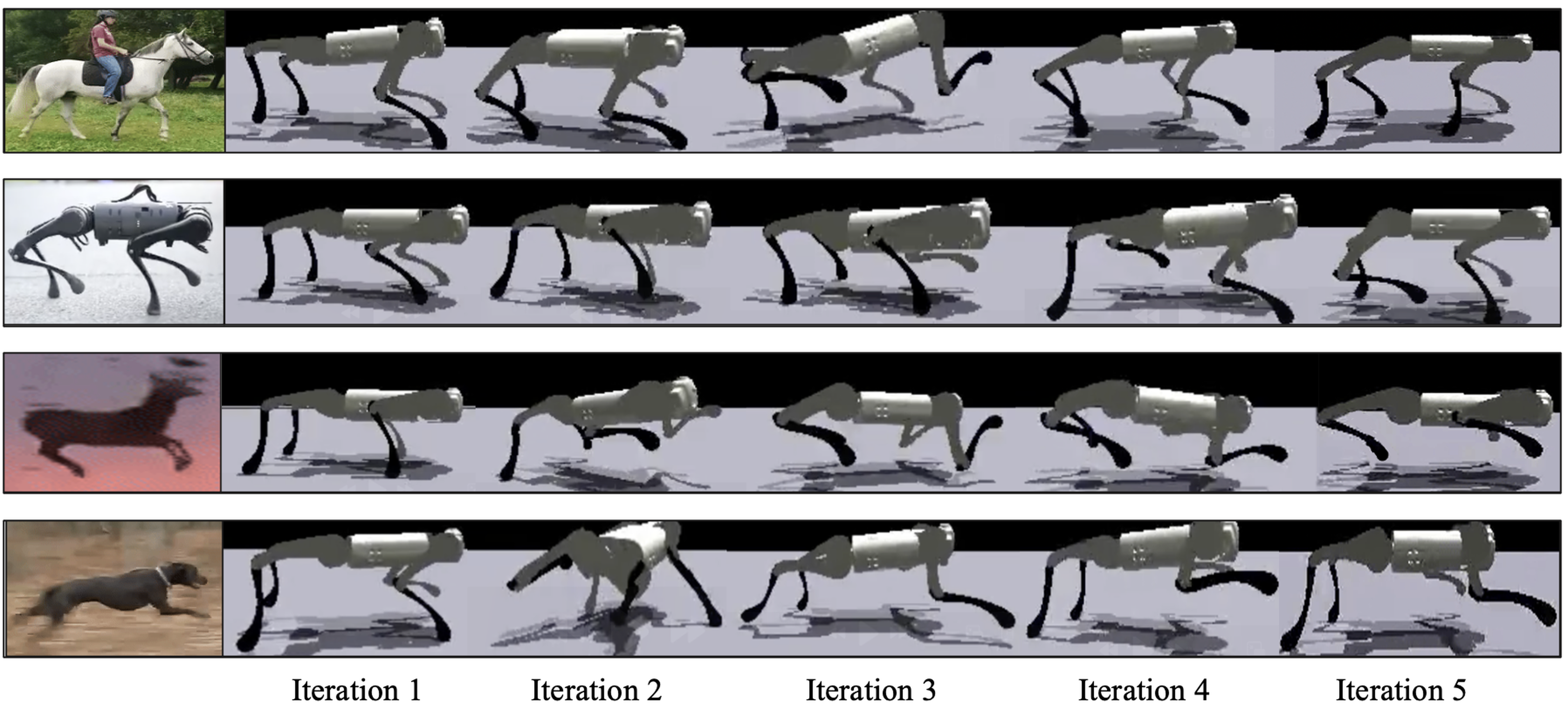}
    \caption{Evolution of task behavior of all skills at a matched gait phase (T=$5$s) across the $5$ SDS reward iterations.}
    \label{fig:ap_photoevol}
\end{figure}

The behavioral evolution of the agents across five iterations of the SDS process, as generated by GPT-4o, illustrates the progression of RF evolution. Snapshots were captured at the same phase of the gait cycle ($T = 5\ \text{s}$) for each iteration, as shown in Fig.~\ref{fig:ap_photoevol}.

\newpage
\subsection{SDS Generalisation}
\label{sec:ap_generalisation}
\begin{figure}[htp!]
    \centering
    \includegraphics[width=\linewidth]{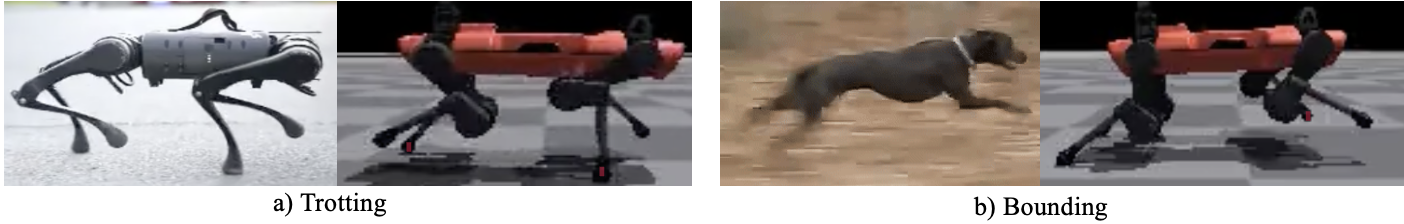}
    \caption{Demonstration of the generalization capabilities of SDS on the ANYmal quadruped robot, featuring an inverted joint configuration and significantly different kinematics. Comparison  between demonstration and learned behavior frames for a)trotting and b) bounding skills.
   (Red corresponds to left-side legs)
    }
    \label{fig:anymal}
\end{figure}

To assess the generalization of SDS, we train the trotting and bounding skills on ANYmal~\citep{anymal}, featuring a different morphology than the one of the demonstrated videos, and therefore altering the kinematic constraints due to its inverted rear knee joints.  
We chose trotting and bounding as the target skills because they are commonly used locomotion behaviors and vary significantly in their dynamic properties. Example imitation pairs are shown in Figure~\ref{fig:anymal}.

\subsection{Comparison Metrics}\label{sec:ap_comaprison_metrics}
We provide details on the metrics selected to compare SDS with state-of-the-art methods of similar scope in Tabel~\ref{tab:metrics_description}, corresponding to Table~\ref{tab:sds_comparison}.

\begin{table}[htp!]
    \centering
    \begin{tabular}{l p{10cm}}
        \toprule
        \textbf{Metric} & \textbf{Description} \\
        \midrule
        Train Time & Duration required to train a single skill on an RTX-4090 GPU, providing a measure of computational efficiency. \\
        DTW & Dynamic Time Warping: a frame-wise imitation accuracy metric measuring temporal alignment between predicted and demonstrated trajectories. Lower is better; $\infty$ denotes poor alignment. \\
        No Human & Whether the method avoids human intervention in reward design, e.g., fine-tuning rewards or fitness functions. A check mark (\cmark) indicates no human-in-the-loop. \\
        No Input Extras & Whether the method avoids requiring additional task inputs like natural language instructions, segmentation masks, or depth maps. \cmark indicates no such inputs are needed. \\
        Real World & Whether the method has been validated in real-world robotic experiments beyond simulation. \cmark denotes successful real-world deployment. \\
        \bottomrule
        \end{tabular}
    \caption{Description of evaluation metrics used for comparing SDS with state-of-the-art methods.}
    \label{tab:metrics_description}
\end{table}

\newpage
\subsection{Real-World Experiments}\label{sec:ap_real_world}
\begin{figure}[htp!]
    \centering
    \includegraphics[width=1\linewidth]{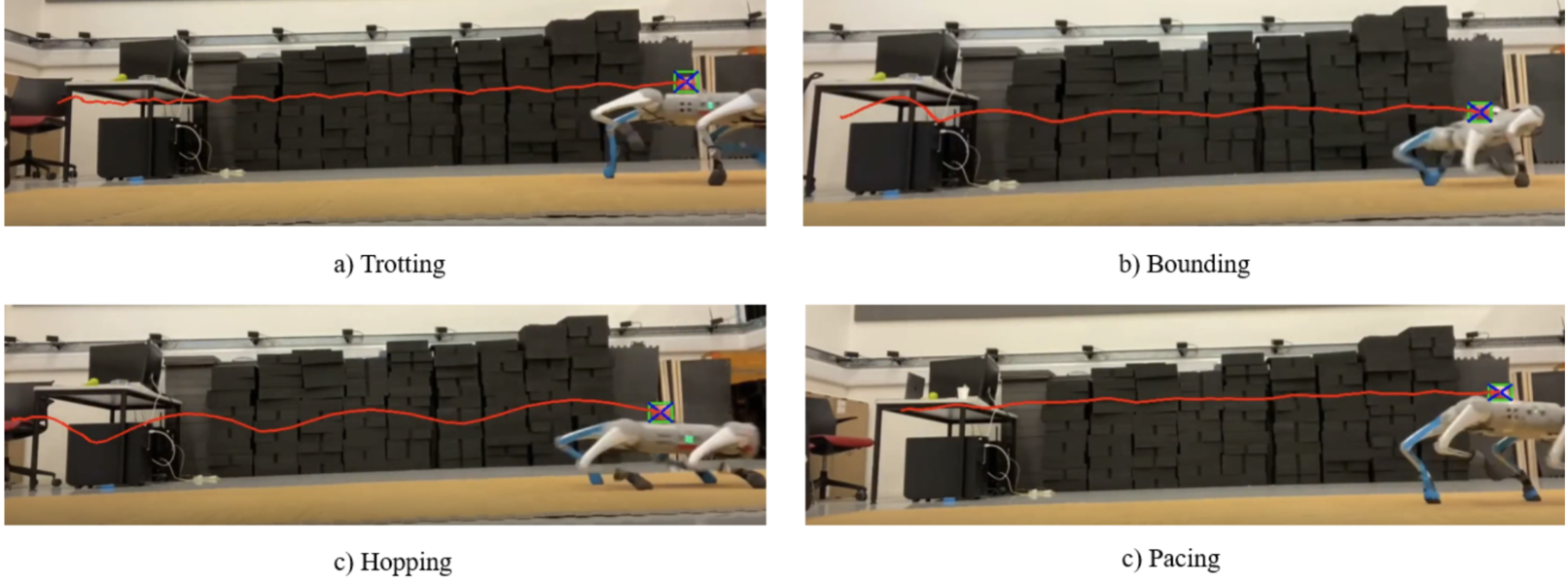}
    \caption{Real Robot Stability: Base Height tracing of real robot, red line indicates the trajectory}
    \label{fig:ap_realworld}
\end{figure}

Showcasing the base height oscillation trajectory traces of the real Unitree Go1 robot across all skills, in Fig.~\ref{fig:ap_realworld}.

\end{document}